\documentclass[11pt]{article}

\usepackage{amsmath,amsfonts,amsthm,amsopn,amssymb}
\usepackage{graphicx}

\usepackage{epsfig,verbatim}

\newtheoremstyle{mythm}{6pt}{6pt}{\itshape}{}{\bfseries}{.}{ }{} 
\newtheoremstyle{mydef}{6pt}{6pt}{}{}{\bfseries}{.}{ }{}         
\newtheoremstyle{myrem}{6pt}{6pt}{}{}{\scshape}{.}{ }{}          

\theoremstyle{mydef}
\newtheorem{definition}{Definition}[section]

\theoremstyle{mythm}
\newtheorem{theorem}[definition]{Theorem}

\newtheorem{corollary}[definition]{Corollary}

\theoremstyle{myrem}
\newtheorem{remark}[definition]{Remark}
\newtheorem{example}[definition]{Example}

\renewcommand{\qedsymbol}{\hfill\rule{2mm}{2mm}}                                                                  

    \DeclareFontFamily{U}{wncy}{}
    \DeclareFontShape{U}{wncy}{m}{n}{<->wncyr10}{}
    \DeclareSymbolFont{mcy}{U}{wncy}{m}{n}
    \DeclareMathSymbol{\sha}{\mathord}{mcy}{"58} 

\newcommand{\R}{\mathbb{R}}

\newcommand{\gs}{\varphi_\sigma}
\def\cF{\mathcal{F}}

\def\cV{\mathcal{V}}

\allowdisplaybreaks

\begin{document}

\begin{center}{ \bf \Large Basic Filters for Convolutional Neural Networks Applied to Music: Training or Design?}\footnote{This work  has been supported by the Vienna Science and Technology Fund (WWTF)
through project MA14-018. }\\[1cm]
  M.~D\"{o}rfler, T.~Grill, R. Bammer, A.~Flexer\\
monika.doerfler@univie.ac.at
\end{center}
Abstract: 
When convolutional neural networks are used to tackle learning problems based on music or other time series, raw one-dimensional data are commonly pre-processed to obtain spectrogram or mel-spectrogram coefficients, which are then used as input to the actual neural network. In this contribution, we investigate, both theoretically and experimentally, the influence of this pre-processing step on the network's performance and pose the question whether replacing it by applying adaptive or learned filters directly to the raw data can improve learning success. The theoretical results show that approximately reproducing mel-spectrogram coefficients by applying adaptive filters and subsequent time-averaging on the squared amplitudes is in principle possible.
We also conducted extensive experimental work on the task of singing voice detection in music. The results of these experiments 
 show that for classification based on Convolutional Neural Networks the features obtained from adaptive filter banks followed by  time-averaging the squared modulus of the filters' output perform better than the canonical Fourier-transform-based mel-spectrogram coefficients.
Alternative adaptive approaches with center frequencies or time-averaging lengths learned from training data perform equally well.

\newpage

\section{Introduction}Convolutional neural networks, first introduced in learning tasks for image data \cite{Lecun98}, have revolutionized state-of-the-art results in many machine learning (ML) problems. 
In convolutional neural networks (CNNs), when  applied to image data, all filter coefficients are usually learned. For  applications to time series, such as audio data, on the other hand, it is common practice to first apply  a fixed filter bank to the raw, one-dimensional data in order to generate a feature representation. In traditional audio signal processing methods, used, e.g., in music information retrieval (MIR) or speech processing, FFT-based features such as \emph{mel-spectrograms} are typically used as such inputs.
These first level features are two-dimensional arrays, derived from some kind of windowed Fourier transform with subsequent mel-scale averaging. 

Recently, the  natural  question arose, what kind of filters a network would learn if it was given the raw audio input.  To date, encouraging  results are scarce and so far, a true end-to-end approach for music signals, i.e., acting on raw audio without any pre-processing, has not been able to outperform models based on linear-frequency spectrogram or mel-spectrogram input \cite{disc14}. It has been argued that these two ubiquitous representations automatically capture invariances which are of importance for {\it all} audio signals, in particular, a kind of translation invariance in time (guaranteed by introducing the non-linear magnitude operation) and a certain stability, introduced by the mel-averaging, to frequency shifts and time-warping (cp.~\cite{anma14}).


In this contribution, we give a formal description of the action of mel-scale averaging on spectrogram coefficients. 
We show that the resulting mel-spectrogram coefficients can indeed be mimicked by applying {\it frequency-adaptive} filters, 
however,  {\it followed by time-averaging}  of each filter's squared amplitude output. In order to obtain  a close approximation to 
mel-spectrogram coefficients, the frequency adaptive filter bank's squared output  signals must each undergo a time-averaging 
operation and the time-averaging window is different for each channel. Furthermore, only dense sampling of the short-time Fourier transform (STFT) leads to almost perfect approximation.  Note that the 
similarity of mel-spectrogram coefficients to the result of time-averaging wavelet coefficients has already been observed in 
\cite{anma14}, without giving a  precise formulation of the connection.\footnote{This observation seems to have served as one 
motivation to introduce the so-called scattering transform, which consists of repeated composition of convolution, a nonlinearity in
 the form of taking the absolute value and time-averaging. In that framework, mel-spectrogram coefficients are interpreted as first 
 order scattering coefficients.} 

 We derive the necessary conditions on the filters, a different one for each bin in the mel-scale, by using the theory  of Gabor 
 multipliers and their spreading function, cf.~\cite{doto10}. Considering the description of an operator by means of its spreading 
 function gives interesting insight in the nature of the correlations invoked by the application of the corresponding operator on the
  signal coefficients. In the case of mel-spectrogram coefficients it turns out that applying wide triangular windows in the high 
  frequency regions actually corresponds to the application of an operator with little spreading in time. This seems to be the 
  intuitively correct choice for audio signals such as music and speech. While a similar effect can be realized by applying wavelet 
  or constant-Q type filters, the subsequent time-averaging alleviates the significant  frequency-spreading effect introduced by 
  rather narrow filtering windows. 
The observation gained from investigating the classical mel-spectrogram  coefficients is thus, that time- and frequency-averaging 
spectrogram coefficients provides  invariances which are useful in most audio classification tasks, cf.~\cite{an13}. On the other 
hand, relaxing the strict averaging performed by computing mel-spectrogram  coefficients may intuitively open the opportunity to 
keep information on details which may be necessary in certain learning tasks. 

In our numerical experiments we thus strive to understand how time and frequency averaging influence CNN prediction performance on realistic data sets. The observations drawn from the experiments on learning filters can be summarized as follows:
\begin{itemize}
\item Using mel-spectrogram coefficients derived from convolutions with a small subsampling factor leads to improved results compared to the canonical FFT-based mel-spectrograms, due to more beneficial influence of the time-frequency subsampling parameters. 
\item Allowing the net to learn center frequencies or time-averaging lengths from the training data leads to comparable improved prediction results.
\item Tricks are required to make the CNNs adapt the feature processing stage at all. Otherwise, the classification part of the network takes over the adaptation required to minimize the target loss.
\end{itemize}


This paper is organized as follows. In the next section we  introduce necessary concepts from time-frequency analysis. In 
Section~\ref{Sec:StrucCNN}, we give a formal description of the network architecture, since we haven't found any concise 
exposition in the literature. Section~\ref{Sec:MelFil} then gives the formal result linking mel-spectrogram coefficients with adaptive 
filter banks. In Section~\ref{Sec:Exp} we report on the experiments with a real-world data set for the problem of singing voice detection. Finally we conclude with a discussion and perspectives in Section~\ref{Sec:Disc}.

\section{Time-frequency concepts}
The Fourier transformation of a function $f \in \mathcal{H}, $ for some Hilbertspace $\mathcal{H}, $ will be denoted by $\mathcal{F}(f).$  We use the normalization 
$\mathcal{F}(f)(\omega) = \int_\mathbb{R} f(t) e^{-2\pi i \omega t} dt$
and denote its  inverse  by  $\mathcal{F}^{-1}(f)(t) = \int_\mathbb{R} f(\omega) e^{2\pi i \omega t} d\omega.$
For  $x, \omega \in \mathbb{R},$
 the translation or time shift operator of a function $f$ is defined as
$$T_xf(t) = f(t-x).$$ 
 and the modulation or  frequency shift operator  of a function $f$ is defined as 
$$M_\omega f(t) = e^{2\pi i t \omega }f(t).$$

The operators of the form $T_xM_\omega$ or $M_\omega T_x$ are called time-frequency shifts.
To obtain local information about the frequency spectrum we define 
 the STFT of a function $f$ with respect to a window $g\neq 0,$ where $f,g \in \mathcal{H}, $ as
\begin{equation}\label{eqSTFT}
\cV_g f (b,k) = \int_t f(t) \overline{g(t-b)} e^{-2\pi i k t} dt =  \mathcal{F} (f\cdot T_b g)(k).
\end{equation}
The STFT can be written as an inner product combining the above operators $$\cV_g f (b,k) = \langle f, M_k T_b g \rangle.$$
Taking the absolute value squared we obtain  the spectrogram as $S_0 (b,k) = | \cV_g f (b,k) |^2$ and $\cV_g g$ is called 
ambiguity function of $g$, reflecting the time-frequency concentration of $g$.  In practice, subsampled and finite versions  of the STFT \eqref{eqSTFT}  are used, cf.~\cite{do01}. Sub-sampling obviously corresponds to choosing certain parts of the available information and this choice can have influence in particular for subsequent processing steps, as we will see in  Section~\ref{Sec:MelFil}.

\section{The structure of CNNs}\label{Sec:StrucCNN}The basic, modular structure of CNNs has often been described, see e.g.,~\cite{Goodfellow-et-al-2016}. Here, we will give a formal statement of the specific architecture used in the experiments in this paper. This architecture has been successfully applied to several MIR tasks and seems to have a prototypical character for audio applications, cf.~\cite{grsc15}.

The most basic building block in a general neural network may be written as
\begin{equation*}
x_{n+1} = \sigma (A_n x_{n}+b_n ) 
\end{equation*}
where $x_{n}$ is the data vector, or array, in the n-th layer, $A_n$ represents a linear operator, $b_n$ is a vector of   biases in the n-th layer and  the nonlinearity $\sigma$ is applied component-wise. Note that in each layer the array $x_n$ may have a different dimension. Now, in the case of convolutional layers of CNNs, the matrix $A$ has a particular structure for the convolutional layers, namely, it is a block-Toeplitz matrix, or, depending on the implementation of the filters, a concatenation of circular matrices, each representing one convolution kernel. There may be an arbitrarily high number of convolutional layers, followed by a certain number of so-called dense layers, for which $A_n$ is again an arbitrary linear operator. In this paper, the chosen architecture comprises up to four convolutional and two or three dense layers. 
\begin{remark}It  has been observed in \cite{mallat, wigrbo17, Wa17,Wa15} that in the context of scattering networks, most of the input signal's energy is contained in the output of the first two convolutional layers. While the context and the filters here are different, this observation might be interesting also as a background for the usual choice of architecture of CNNs for audio processing.  
\end{remark}

\subsection{The CNN with spectrogram input}\label{Sec:CNN_Spec}
The standard input in learning methods for audio signal  is based on a sub-sampled spectrogram, either in its raw form, or after some pre-processing such as the computation of mel-spectrogram, defined in \eqref{eq:melspec}, which we will consider in detail in Section \ref{Sec:MelFil}. In any case, the input to the CNN is an array of size $M\times N$. 
\begin{remark}In most MIR tasks, the inputs are derived from rather short snippets, that is, about 2 to 4 seconds of sound. Considering a sampling rate of 22050\,Hz, a window size of $2048$ samples and a time shift parameter of $512$ samples, i.e., 23\,ms, the resulting spectrogram (containing positive frequencies only) is of size   $M\times N = 1024\times 130$, where the latter is the time dimension. Hence, the frequency dimension is, in some sense, over-sampled. In particular, individual bins in the higher frequency regions contain less energy and thus information than in lower regions.  Computing the mel-spectrogram is a convenient and straight-forward method of reducing the information to typically $80$ frequency channels by averaging over increasingly many frequency bins.
The number of $80$ channels has been determined with preliminary experiments as a breakpoint for optimal CNN performance, obviously because of a sufficient resolution along the frequency dimension. The same setting has already been used in the reference implementation \cite{schlueter:2015-ismir} for the experiments of Section~\ref{Sec:Exp}.
\end{remark}
We now define the following building blocks of a typical CNN: 
\begin{itemize}
\item Convolution:  $S\ast w (m,n ) := \sum_{m'} \sum_{n'} S(m',n') w(m-m', n-n')$
\item Pooling: For $1\leq p \leq\infty$, we define $A\times B$ pooling as the operator mapping an $M\times N$ array $S_0$ to a $M/A\times N/B$ array $S_1$ by 
\begin{equation*}
S_1 (m,n ) = P_p^{A,B} (m,n) = \| v_{S_0}^{m,n}\|_p
\end{equation*}
where $v_{S_0}^{m,n}$, for $m  = 1, \ldots , M/A$ and $n = 1, \ldots , N/B$,  is  the vector consisting of the array entries $S_0((m-1)\cdot A+1,\ldots, m\cdot A;(n-1)\cdot B+1,\ldots, n\cdot B)$. In this work, we use max-pooling, which has been the most successful choice, corresponding to $p = \infty$ in the above formula.
\item A nonlinearity $\sigma : \mathbb{R}\mapsto  \mathbb{R}$, whose action is always to be understood component-wise. In all but the last layer we use leaky rectified linear units, which allow for  a small but non-zero gradient when the unit is not active:
\[ \sigma (x)={\begin{cases}x&{\mbox{if }}x>0\\c x&{\mbox{otherwise}}\end{cases}}\] for some $c\ll 1$. The output layer's nonlinearity $\sigma_o$ is a sigmoid function. 
\end{itemize} 
We now denote  the input array to a convolutional layer by $S_{n}\in \R^{M_n\times N_n\times K_n}$, where $K_n$ is the number of feature maps of size $M_n\times N_n$ in layer $n$, i.e.,  $S_{n}(k_n)\in \R^{M_n\times N_n}$ for $k_n = 1,\ldots , K_n$.
Using the above definitions,
we can now write the output of (convolutional)  layer $n+1$ with convolutional kernels $w_{n+1} \in \R^{K_{n+1}\times K_n \times M_n\times N_n}$ as follows: 
\begin{equation}\label{OneLayerOutput}
S_{n+1}(k_{n+1})\! = \!  P^{A_n,B_n}_{\infty} \sigma\left[ \left(\! \sum_{k_{n}=1}^{K_{n}}S_{n}(k_{n})\ast w_{n+1}(k_{n+1}, k_n)\right) \! +b^{k_{n+1}}\otimes\bf{1}\right]
\end{equation}
where $\bf{1}$  is an all-ones array of size  $M_n\times N_n$, $b^{k_{n+1}}\in\R^{K_{n+1}}$ and $S_{n+1}(k_{n+1})\in \R^{M_n/A_n\times N_n/B_n}$ for $k_{n+1} = 1, \ldots , K_{n+1}$.\\
To formally describe the final, dense layers, we let $D_c$ denote the number of convolutional layers and $S_{D_c}\in \R^{M_{D_c}\times N_{D_c}\times K_{{D_c}}}$ the output of the last convolutional layer. Then the over-all action of a CNN with two dense layers and a single output unit emitting $x_{out}$, can be written as
\begin{equation}\label{AllOutput}
x_{out} = \sigma_o (\mathcal{A}_2\cdot  [\sigma (\mathcal{A}_1 \cdot S_{D_c}+b_{D_c+1} )]+b_{D_c+2}).
\end{equation}Here, $\mathcal{A}_1$ and $\mathcal{A}_2$ are weight-matrices of size $N_d \times M_{D_c} N_{D_c}K_{{D_c}}  $ and $1 \times N_d $, respectively, where $N_d$ is the number of hidden units in the first dense layer, $b^{D_c+1} \in \R^{N_d}$ and 
$b^{D_c+2} \in \R$.

\subsection{Modifying the input array}\label{Sec:ModIn}
As mentioned in the previous section, the spectrogram of audio is often pre-processed in order to reduce the dimensionality on the one hand, and in order to obtain a spectral representation that better fits both human perception and properties of speech and music on the other hand. Additionally, the authors in \cite{anma14} pointed out that using mel-spectrogram instead of the spectrogram guarantees improved stability with respect to frequency shifts or, more generally, deformations of the original audio signals, than the usage of spectrograms.    However, {\it given appropriate choice of  network architecture}, comparable results can usually be achieved using either the spectrogram or the mel-spectrogram, i.e., the invariance introduced by the mel-averaging can also be learned.  In other respects, omitting the frequency-averaging provided by the mel-spectrogram leads to an increase in the number of weights to be learned. On the other hand, these observations raise the question, whether using filters learned directly in the time-domain, would improve the net's ability to achieve the  amount of invariance most appropriate for a particular ML task and thus increase stability. The corresponding approach then implies learning time-domain filters  already in a layer prior to the first $2D$-convolution. To put this remark into perspective, we note that the spectrogram may easily be interpreted as the combined (and possibly sub-sampled) output of several convolutions, since, setting $ h\check{}(n) = h(-n)$, we can write
\begin{equation*}
S_0 (m,n) = | \sum_{n'} f(n') h(n'-n) e^{-2 \pi i m n'}|^2 =
|f\ast h\check{}_m (n)|^2
\end{equation*}

\subsection{Questions}\label{Sec:Questions}
In the two following sections we thus raise and answer two questions:
\begin{enumerate}
\item[(i)]  Is it possible to obtain coefficients which are approximately equivalent to the well-established mel-spectrogram coefficients simply by using the `correct' filters directly on the audio signal?
\item[(ii)] Can  adaptivity in frequency- and time-averaging  improve prediction accuracy? In particular, for a given set of frequency-adaptive filters precisely mimicking the mel-scale, can  a time-averaging layer with learned averaging width improve learning performance?
\end{enumerate}

\section{The mel-spectrogram and basic filters}\label{Sec:MelFil}
In this section, we take a detailed look at the mel-spectrogram. This representation is derived from the classical spectrogram by weighted averaging of the absolute values squared of the STFT and can 	undoubtedly be referred to as the most important feature set used in speech and audio processing, together with MFCCs which are directly derived from it.
The number of mel-filters used varies between $80$ filters between $80$ Hz and $16$ kHz~\cite{grsc15} and $128$~\cite{disc14} or more. In order to better understand the relation between the result of mel-averaging and FFT-based analysis with flexible windows, we observe the following: denote the input signal by $f\in\mathbb{C}^N,$ the window function for generating the spectrogram by $g\in\mathbb{C}^N$ and the mel-filters, typically given by simple triangular functions,  by $\Lambda_\nu \in\mathbb{C}^N$ for $\nu \in \mathcal{I} = \{1,\ldots ,K\}$, where $K$ is the chosen number of filters. We can then write the mel-spectrogram as

\begin{equation}\label{eq:melspec}
\operatorname{MS}_{g}(f) (b,\nu ) = \sum_k |\mathcal{F} (f\cdot T_b g)(k)|^2 \cdot \Lambda_\nu (k).
\end{equation}
And\'en and Mallat postulated  in~\cite{anma14}, that the mel-spectrogram can be approximated by time-averaging the absolute values squared of a wavelet transform. Here, we make their considerations precise by showing that we can get a close approximation of the mel-spectrogram coefficients if we use adaptive filters. 
\begin{remark}
Note that the resulting transform may be interpreted as a nonstationary  Gabor transform, compare~\cite{badohojave11,badokoto13,dogrhove13,badogr17}.  \end{remark}
In practice one always uses a sub-sampled version of the STFT, i.e., we consider time-sampling points in $\alpha \mathbb{Z}$ and the Fourier transform in \eqref{eq:melspec} is sampled on $\beta \mathbb{Z}$. 
We then compare two different settings which lead to a time-frequency feature map which is then used as input to the deeper layers of the CNN: 
\begin{enumerate}
\item STFT-based: Compute spectrogram and take weighted averages over certain regions in frequency; for the classical mel scale this leads to the mel-spectrogram coefficients, but other choices of $\Lambda_\nu$ are possible. Taking time- and frequency-sampling parameters $\alpha, \beta$ into account, the resulting  time-frequency feature map is computed for $b = \alpha  l_0$ as follows
\begin{equation}\label{eq:FSdef}
\operatorname{MS	}_{g}(f) (b,\nu ) = \sum_k |\mathcal{F} (f\cdot T_{b} g)(\beta k)|^2 \cdot \Lambda_\nu (\beta k).
\end{equation}
\item Filter bank-based: compute filtered version of $f$ with respect to some, possibly adaptive, filter bank $h_\nu$, $\nu\in\mathcal{I}$ and apply subsequent time-averaging using a time-averaging function $\varpi_\nu$:
\begin{equation}\label{eq:FFBdef}
\operatorname{FB}_{h_\nu}(f) (b,\nu ) =  \sum_l |(f\ast h_\nu)(\alpha l)|^2 \cdot \varpi_\nu(\alpha l-b).
\end{equation}
\end{enumerate}
The following central theorem gives an estimate for the difference between the two above approaches for each entry in the feature maps. 

\begin{theorem}\label{Th:subsampled case}For all $\nu \in \mathcal{I}$, let  $g,h_\nu, \Lambda_\nu, \varpi_\nu$ be given. Let 
$\operatorname{MS}_{g}(f)$ and $\operatorname{FB}_{h_\nu}(f)$ be computed  on a lattice $\alpha \mathbb{Z}\times \beta \mathbb{Z}$ and set
\begin{equation}\label{eq:defM}
\mathcal{M}^\nu (x) = \sum_l T_{\frac{l}{\beta} }\mathcal{F}^{-1} (\Lambda_\nu ) (x)\, \mbox{ and }\,
 \mathcal{M}^\nu_F(\xi) = \sum_k T_{\frac{k}{\alpha } }\mathcal{F}(\varpi_\nu) (\xi). 
\end{equation}
Then the following estimate holds for all $(b, \nu)\in  \alpha \mathbb{Z}\times \mathcal{I}$:
\begin{eqnarray}\label{eq:errboundaliases}
|\operatorname{MS}_{g}(f)(b,\nu )-\operatorname{FB}_{h_\nu}(f)(b,\nu )| 
\leq\| \cV_g g \cdot \mathcal{M}^\nu  - \cV_{h_\nu} h_\nu \cdot  \mathcal{M}^\nu_F \|_2 \cdot \|f\|_2^2
\end{eqnarray}
\end{theorem}
A technical proof of Theorem~\ref{Th:subsampled case} is included in Appendix~\ref{AppA}. The basic idea of the proof lies in expressing both 
$\operatorname{MS}_{g}(f)$ and $\operatorname{FB}_{h_\nu}(f)$ by means of a bilinear form generated by different specific time-frequency 
multipliers. The underlying operators can then be compared using their respective {\it spreading functions}, \cite{doto10,feko98}, 
 an alternative operator description. An operator's spreading function gives an intuition about the operator's action in the space 
 of time-lag and frequency-lag.  Time-frequency multipliers' spreading functions enjoy a simple form,  which is simply the product 
 of the analysis windows' ambiguity function with a two-dimensional Fourier transform of the multiplier sequence. Figure~\ref{fig:3} shows the ambiguity functions  $\cV_g g(x,\xi )$, $\cV_{h_\nu} h_\nu (x,\xi )$ which would correspond to the operators without frequency- or time-averaging, respectively, and the weighted ambiguity functions 
   $\cV_g g (x,\xi )\cdot \mathcal{F}^{-1} (\Lambda_\nu ) (x)$,  corresponding to the ambiguity function after mel-averaging in frequency by $\Lambda_\nu$ and $\cV_{h_\nu} h_\nu(x,\xi )\cdot \mathcal{F} (\varpi_\nu )(\xi)$ corresponding to the filter bank approach after time-averaging by $\varpi_\nu $. It is obvious that frequency-averaging reduces the time-lag of the operator while time-averaging reduces the higher frequency-lag introduced by the narrower windows in the adaptive filter bank; over-all very close behavior can be achieved with both approaches, in particular with small $\alpha, \beta$.  
For the fully sampled case, i.e., $\alpha = \beta = 1$, we obtain the following expression:
\begin{eqnarray}\label{eq:errbound}
\|\operatorname{MS}_{g}(f) -\operatorname{FB}_{h_\nu}(f)  \|_\infty \leq 
\| \cV_{h_\nu} h_\nu  \cdot \mathcal{F} (\varpi_\nu ) - &\cV_g g \! \cdot\! \mathcal{F}^{-1} (\Lambda_\nu ) \|_2\! \cdot\! \|f\|_2^2
\end{eqnarray}
This leads to a statement about precise recovery of mel-spectrogram coefficients by filter bank approximation. 
\begin{remark}
A preliminary version of the following statement has been announced without proof in \cite{badogr17}. \end{remark}
\begin{corollary}\label{cor:timeavg}
Let an analysis window $g$ and  mel-filters $\Lambda_\nu $ be given, for $\nu\in \mathcal{I}$.  
If, for each $\nu$,  the windows $h_\nu$ and {\it time-averaging functions} 
 $\varpi_{\nu}$ are chosen such that
\begin{equation}\label{eq:melcond}
\cV_{h_\nu} h_\nu (x,\xi ) \cdot \mathcal{F} (\varpi_\nu )(\xi) = \cV_g g (x,\xi ) \cdot \mathcal{F}^{-1} (\Lambda_\nu ) (x) ,
\end{equation}
then the mel-spectrogram coefficients  can be obtained by time-averaging the filtered signal's absolute value squared, i.e., 
for all $(b, \nu)\in   \mathbb{Z}\times \mathcal{I}$:
\begin{equation}\label{eq:TA}
\operatorname{MS}_{g}(f) (b,\nu ) = \operatorname{FB}_{h_\nu}(f) (b,\nu ) .
\end{equation}
\end{corollary}
\begin{example}
While it is in general tedious to explicitly derive conditions for the optimal filters $h_\nu$ and the time-averaging windows  $\varpi_{\nu}$, we obtain a more accessible situation if we restrict the choice of windows to dilated Gaussians $g(t)  = \gs (t) = (\frac{2}{\sigma})^{ \frac{1}{4}} e^{-\pi  \frac{t^2}{\sigma}}$, for which $\cV_{\gs} \gs (x,\xi ) = e^{-\frac{\pi}{2}  \frac{x^2}{\sigma}}e^{-\frac{\pi}{2}\sigma \xi^2 }e^{-\pi  i x \xi}$. Thus,  
fixing  $g = \gs $  for some scaling factor $\sigma$, 
letting  the filters $\Lambda_\nu $ be given as shifted and  dilated versions of a basic shape (e.g., in the case of mel-filters, asymmetric triangular functions), i.e., $\Lambda_\nu (\xi ) = T_\nu D_{a(\nu)}\Lambda(\xi )$, for $\nu\in \mathcal{I}$ and
assuming that each filter $h_\nu$ is  a dilated and modulated Gaussian window i.e., $h_\nu ( t ) = e^{2\pi i \nu t}\varphi_{\rho (\nu )} (t) $, 
condition \eqref{eq:melcond} leads to the following conditions in separate variables: 
\[
e^{-\frac{\pi}{2}  x^2(\frac{1}{\rho (\nu )}- \frac{1}{\sigma})} =  \mathcal{F}^{-1} (D_{a(\nu)}\Lambda) (x)  \mbox{ and }
e^{-\frac{\pi}{2} \xi^2 (\sigma-\rho (\nu )) }   = \mathcal{F} (\varpi_\nu)(\xi).
\]
\end{example}
From the example it can be seen that even in the case of Gaussian analysis windows a precise recovery of standard mel-spectrogram coefficients is possible, 
if the involved analysis windows and averaging windows are appropriately scaled Gaussians. In the more realistic case of compactly supported 
analysis windows such as Hann windows, triangular frequency-averaging functions $\Lambda_\nu$ typically used for computing the mel-spectrogram coefficients and coarser sampling schemes, we have to resort to alternative methods for obtaining the filter bank-based approximation. 

\subsection{Computation and examples of  adaptive filters}
We now describe the strategy for computing  adaptive filters  leading to a filter bank-based approximation of mel-like coefficients 
based on general windows. Very often, these windows will be compactly supported and their STFT will not factorize in two components in separate variables  $x,\xi$. Since $g$ and $\Lambda_\nu$ are fixed, and $\mathcal{F}(\varpi_\nu ) (\xi)$ only allows for a multiplicative constant in each frequency bin, we can only perfectly adapt $h_\nu$ in one frequency bin. We thus use the following trick for computing $h_\nu$ for a given mel-filter $\Lambda_\nu$: we consider the right-hand side of 
 \eqref{eq:melcond} in $\xi = 0$. This is justified  if  $\mathcal{F} (g)$, and thus $|\cV_g g(x, \xi)| \leq( |\hat{g}|\ast |\hat{g}|)(\xi) $, decays fast in the frequency variable $\xi$; this is typically the case for the windows used in practice, such as Hann windows, since we strive to obtain separation between the frequency-bins. Therefore, the component at $x=0$ will have by far the strongest influence on the error made when minimizing \eqref{eq:errbound} and we will use it to obtain $h_\nu$. We then have, with $\check{g} (x) =  \overline{g}(-x)$, the following version of  \eqref{eq:melcond}: 
 \begin{eqnarray*}
\cV_{h_\nu} h_\nu (x,0) \cdot \mathcal{F}(\varpi_\nu ) (0) =&\cV_g g (x,0) \cdot \mathcal{F}^{-1} (\Lambda_\nu ) (x) \\
\Rightarrow(h_\nu\ast \check{h_\nu} ) (x) \cdot\mathcal{F}(\varpi_\nu ) (0) =&(g\ast \check{g} ) (x) \cdot \mathcal{F}^{-1} (\Lambda_\nu ) (x).\end{eqnarray*} 
Taking  Fourier transform on both sides  we obtain
 \begin{eqnarray*}
\mathcal{F} (h_\nu\ast \check{h_\nu)}  = |\mathcal{F}(h_\nu)(\xi)|^2= & \mathcal{F} ((g\ast \check{g} ) \cdot \mathcal{F}^{-1} (\Lambda_\nu ) )(\xi ),
\end{eqnarray*} and compute $h_\nu$ as 
 \begin{eqnarray*}
h_\nu (t) = & \mathcal{F}^{-1} \left(\sqrt{\mathcal{F} ((g\ast \check{g} ) \cdot \mathcal{F}^{-1} (\Lambda_\nu ) )}\right)(t)
\end{eqnarray*} 
Similarly, by setting $x = 0$ in the left-hand side of \eqref{eq:melcond}, we compute 
\begin{equation*}
\mathcal{F} (\varpi_\nu )(\xi) = \cV_g g (0,\xi ) \cdot \mathcal{F}^{-1} (\Lambda_\nu ) (0) /\cV_{h_\nu} h_\nu (0,\xi )
\end{equation*} where we only consider values of $\cV_{h_\nu} h_\nu (0,\xi )$ above a threshold $\varepsilon$.

\begin{figure}[!]
  \centering
     \includegraphics[width=0.9\columnwidth]{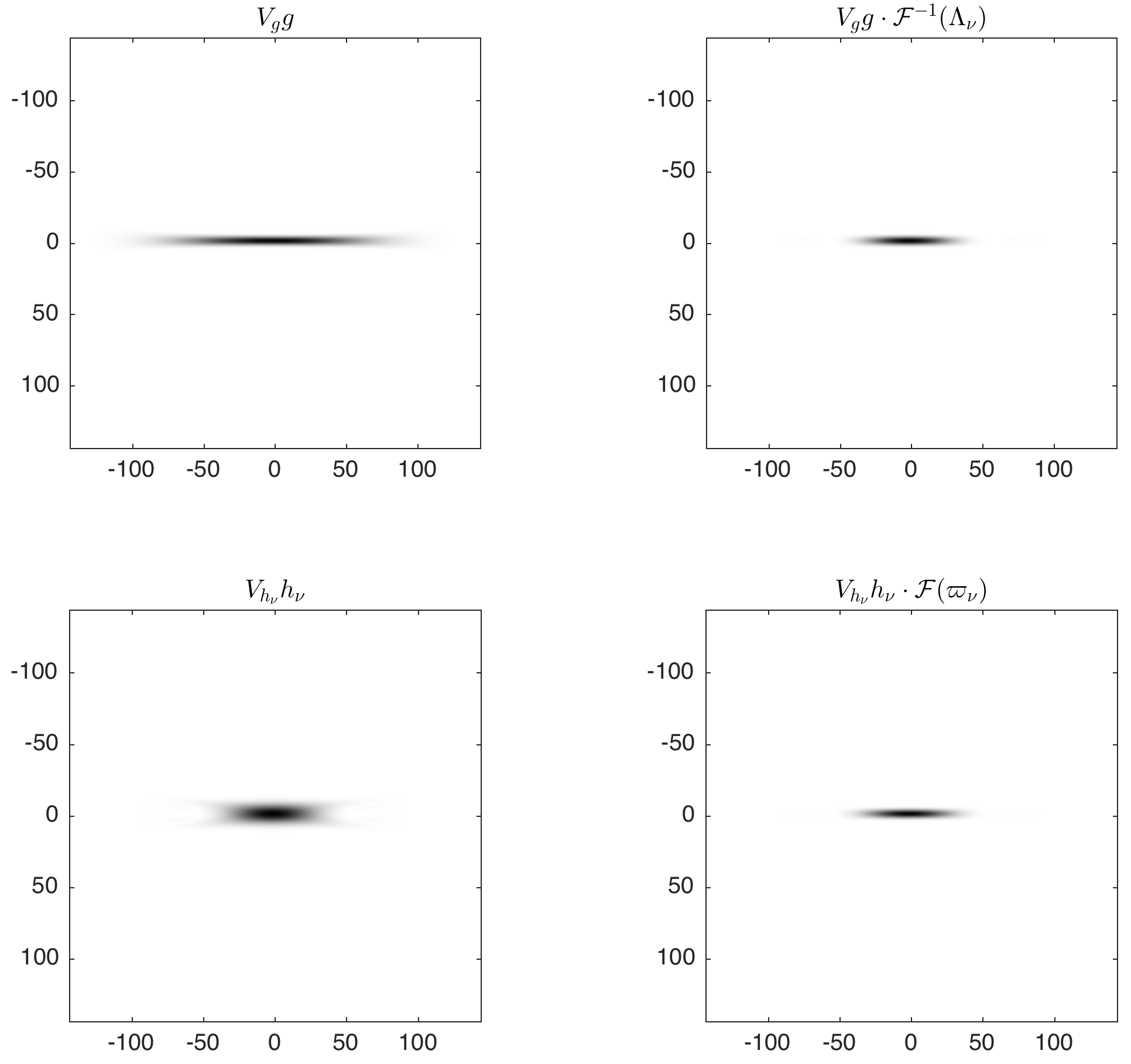}
  \caption{Ambiguity functions  $\cV_g g$, $\cV_{h_\nu} h_\nu$,  and  weighted ambiguity functions 
   $\cV\cV_gg g \cdot \mathcal{F}^{-1} (\Lambda_\nu )$, $\cV_{h_\nu} h_\nu\cdot \mathcal{F} (\varpi_\nu )(\xi)$ used for the computation of adaptive filtering,  for $\nu = 50$.  It is clearly visible that the surplus in frequency spread introduced by the narrower window $h_\nu$ is removed by time-averaging. On the other hand, frequency averaging reduces the time-spread of the wider window $g$.}
  \label{fig:3}
\end{figure}
We now give some examples of filters $h_\nu$ computed to obtain mel-coefficients $\operatorname{MS}_{g}(f) $ by 
time-averaging $ |(f\ast h_\nu)(l)|^2$ as in \eqref{eq:TA} following the procedure described above. We consider  Hann windows, which is the standard choice in audio 
processing,  also applied in the computation of  mel-spectrogram coefficients and their approximation in Section~\ref{Sec:Exp}.
 Starting from a Hann window $g$, we  compute adaptive filters $h_\nu$ for  $80$ bins of the mel-scale. 
Figure~\ref{fig:3} shows the ambiguity functions  $\cV_g g$, $\cV_{h_\nu} h_\nu$,  and the weighted ambiguity functions 
   $\cV_g g \cdot \mathcal{F}^{-1} (\Lambda_\nu )$, $\cV_{h_\nu} h_\nu\cdot \mathcal{F} (\varpi_\nu )(\xi)$, for $\nu = 49$, which corresponds to 2587.6\,Hz. 
   
\begin{figure}
  \centering
     \includegraphics[width=\columnwidth]{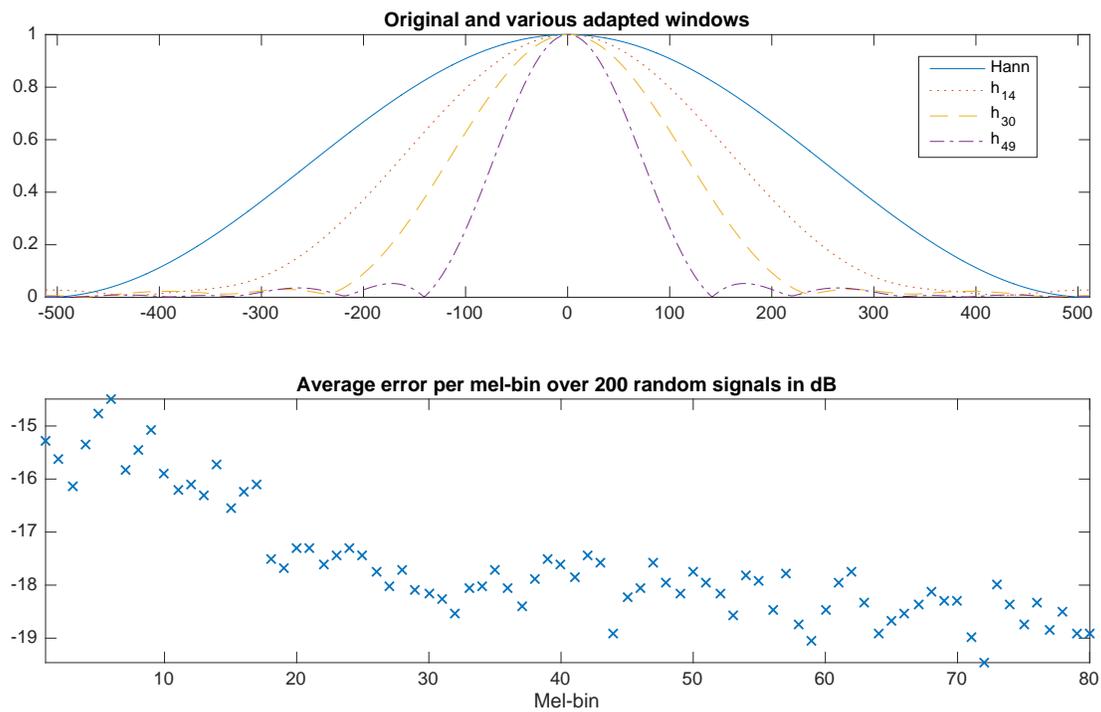}
  \caption{Upper plot: (original) Hann window and adapted windows; lower plot: error in approximation of mel-spectrogram coefficients by adaptive filtering and subsequent time-averaging on the squared amplitudes}
  \label{fig:1}
\end{figure}

In Figure~\ref{fig:1}, the upper plot shows the original Hann window $g$, which had been used to compute the spectrogram from which the mel-spectrogram coefficients are derived, and three adapted windows. Note that the adapted windows get shorter in time  with increasing  mel-number; this effect serves 
to realize the mel-averaging by adaptivity in the frequency domain. The lower plot shows the average error per bin obtained 
from computing the mel-spectrogram coefficients and their approximations for 200 (normally distributed) random signals. 

For an illustration of the time-frequency representations applied to a real audio signal, cf.~Figure~\ref{fig:tf_repr}.


\section{Experiments on Singing Voice Detection}\label{Sec:Exp}
In Theorem~\ref{cor:timeavg} it is shown that  coefficients with mel-characteristics (and other related nonlinear scales) 
can be closely approximated by  applying appropriately chosen filters directly to raw audio data and allowing for a subsequent time-averaging step on the squared absolute output.
Now, we are interested in investigating if the theoretical findings translate to typical real-world problems that have
 already been successfully treated with CNNs. Thereby, we are motivated by the fact that state-of-the-art results
  for several MIR problems are based on mel-spectrogram coefficients which show certain desirable invariance and 
  stability properties. 
In particular, due to the modulus, they are invariant to translation and, due to the frequency averaging, they exhibit stability to 
certain deformations such as time-warping, cf.~\cite{anma14}. 
However, in general, the required invariance and stability with respect to deformations will depend on data characteristics and the learning task, cf.~\cite{ma16}.
In our experiments, we hence start from the filter bank-based computation of approximative mel-coefficients, cf.~\eqref{eq:FFBdef}.  In the sequel, the results obtained from the filter bank-based  coefficients can serve as a reference 
point and they 
should   not be  significantly worse than 
the results achieved when using the standard mel-coefficients as input. This reference is necessary, since  certain 
 implementation details are different
for the original mel-coefficients and their filter bank approximation. This concerns, 
in particular,  pre-processing steps such as batch normalization or padding. The adaptation of the time-averaging starts from this 
 implementation,  so that we needed to rule out adversarial effects stemming from sources other than the adaptation process. 
 For the adaptive scenarios,  we allow parts of the feature processing stage (time-averaging lengths, center frequencies) 
 to be learned by the network, posing the question whether adaptivity in this step can improve the network's performance.

We need to note that, when trained on a specific problem, both the feature layers (including the adaptive time-averaging step) 
and the classification part of a CNN will concurrently adapt their parameters towards optimally predicting the given targets.
 We will discuss the implications of this behavior for our experiments in Section~\ref{ssec:exp_results}.

Our hypothesis is that a CNN with an architecture that is adapted to a given learning task will learn filters---in this case their adaptive components---which alleviate the extraction of stabilities and invariance properties and are thus beneficial in the given context.
\subsection{Data}
We investigate the effects of learning filters directly on raw audio by revisiting the problem of \emph{singing voice detection}~\cite{schlueter:2015-ismir}
we have studied before.
In the referenced publication, a CNN was tuned for maximum prediction accuracy both in the absence or presence of various forms of data augmentation.

The experiments were performed on a non-public dataset of 188 30-second audio snippets from an online
music store (dataset `In-House A'), covering a very wide range of genres and origins.
For the evaluation we used a five-fold cross-validation with slightly unequal folds, for each iteration 150 or 151 files for training, the remaining 37 or 38 for evaluation.
The testing folds are non-overlapping and add up to the total of 188 items.
The audio was subsampled to a sampling rate of 22.05\,kHz and down-mixed to mono.
The mel-spectrograms were calculated using an STFT
with Hann windows, a frame length of 1024 and a frame rate of 70 per second (equivalent to a hop size of 315 samples).

For this paper, instead of magnitude spectra, as in the reference model, we use power spectra as in \eqref{eq:melspec}, 
also following the convention used in \cite{anma14}. 
We apply a filter bank with 80 triangular mel-scaled
filters from 27.5\,Hz to 8\,kHz, then logarithmize the squared magnitudes
(after clipping values below $10^{-7}$).
%

We have also left out any form of data augmentation. For the context of this paper, where we are interested in fundamental qualities of feature representation rather than maximum prediction performance, data augmentation would not be beneficial,
but would rather negatively impact training times.

\subsection{CNN training procedure and architecture}\label{Sec:TP}

The training procedure used in our experiments is slightly different than in the reference publication~\cite{schlueter:2015-ismir}.
The networks are trained on mel-spectrogram excerpts of 115 spectrogram frames ($\sim$1.6\;seconds) paired with a binary label denoting the presence or absence of human voice in the central frame. Training is performed using stochastic gradient descent on cross-entropy error based on mini-batches of 64 randomly chosen examples.
Updates to the network weights are computed using the ADAM update rule~\cite{adam15} with an initial learning rate of 0.001 and an adaptive scheme reducing the learning rate twice by a factor of 10 whenever the training error does not improve over three consecutive episodes of 1000 updates.
Evaluation is performed running a complete five-fold cross-validation run to obtain predictions for the whole set of training data, with this procedure repeated multiple times with different network initialization and data ordering.

As described in Section~\ref{Sec:CNN_Spec}, the applied CNN architecture employs three types of feed-forward
neural network layers: convolutional \emph{feature processing layers} convolving
a stack of 2D inputs with a set of learned 2D kernels,
\emph{pooling layers} subsampling a stack of 2D inputs by taking
the maximum over small groups of neighboring pixels, and
dense \emph{classification layers} flattening the input to a vector and applying a
dot product with a learned weight matrix $\mathcal{A}_j$.

The architecture used in~\cite{schlueter:2015-ismir} has a total number of 1.41~million weights, with the dense connections of the classification layers taking up the 
major share (1.28~million, or 91\%).
It can be expected that the actual output of the convolutional feature stage is of subordinate importance
when the classification stage with its high explanatory power dominates the network.

If data augmentation is not considered,  the network size---especially the classification part---can be drastically reduced while largely preserving its performance. 
This size reduction is possible, since, as a general rule, the necessary number of parameters determining the network is correspondent to the complexity of the training data set. As we are interested in the impact of the convolutional feature stage's properties, we reduce the architecture for our experiments as follows: 
We use four convolutional layers, two $3\times3$ convolutions of 32 and
16 kernels, respectively, followed by $3\times3$ non-overlapping
max-pooling and two more $3\times3$ convolutions of 32 and
16 kernels, respectively, and another $3\times3$ pooling stage.

With the conventions of \eqref{OneLayerOutput}, with a slight abuse of notation by noting the number of non-zero elements in the convolutional kernels instead of the underlying dimension of convolution, the applied setting corresponds to 
\begin{itemize}
\item $K_0 =1$,  $K_1 = K_3 = 32$, $K_2 = K_4  = 16$; 
\item $w_1\in \R^{32\times 1\times 3\times 3}$,   $w_2\in \R^{16\times 32 \times 3\times 3}$;
\item $A_1 = B_1 = 1$, $A_2 = B_2 = 3$
\item  $w_3\in \R^{32\times 16 \times 3\times 3\times 16}$,  $w_4\in  \R^{16\times 32 \times 3\times 3}$;
\item $A_3 = B_3 = 1$, $A_4 = B_4 = 3$
\end{itemize}

For the classification part, we experimented with two variants:
One with two dense layers of 64 and 16 units (`small-two'), and the other one with just one dense layer of 32 units (`small-one').
In both cases, the final dense layer is a single sigmoidal output unit.
For the first variant, the total number of weights is 94337, with the classification stage taking up 79969 units, or 85\%.
The second variant features a considerably smaller classification network: the total number of weights is 53857, with the classification stage taking up 39489 units, or 73\%.
The different network sizes, especially the ratio of feature to classification stage allows us to analyze the influence of the different parts. 
Specifically, we expect the performance of the `small-one' architecture to be more directly connected to the quality of the time-frequency representation.


\subsection{Experimental setup}\label{Sec:ExpSetup}

In the following, we will compare the behavior of the CNNs applied to the STFT-based mel-spectrogram features to
features computed using filter banks as described in Section~\ref{Sec:MelFil}. Both are computed in end-to-end fashion ad hoc from the audio signal. 
The maximum kernel sizes of the filter banks are set to 1024, identical to the frame length of the previously used STFT.
The training examples are snippets of the audio signal with a length of $115\times315 + 1024 - 1 = 37248$ samples each with a hop size of 315 samples.

To judge the influence of adaptivity,  four different approaches have been compared: 
\begin{enumerate}
\item\label{item:1} `Filter bank, approximation': Filter bank and time averaging as derived in Section~\ref{Sec:MelFil}
\item\label{item:2} `Filter bank, naive': Filter bank with Hann envelopes. 
The kernel size equals the time support for the lowest frequency band (50\,Hz) and reduces, according to the band-width requirements of the mel frequency scale, down to 94 samples for the highest band at 7740\,Hz.
After the filter bank, fixed-size time-averaging by pooling  for improved computational efficiency.
\item\label{item:3} `Filter bank, fixed-width': Filter bank as in \ref{item:2}., but with fixed-size time-averaging using a convolution with a Hann window
\item\label{item:4} `Filter bank, variable-width': Adaptive time-averaging after the filter bank, with individual adaptation per frequency bin, learned from the training data.
\end{enumerate}

For reasons of computational cost it is not feasible to perform a full sample-by-sample convolution for the filter bank. 
For the bulk of our filter bank experiments, we have chosen a convolution stride for the filters of 21 samples, that is, 
the resulting spectrum is down-sampled along the time axis by a factor of 21.
The subsequent non-overlapping averaging is computed on 15 frames each, in order to stay comparable with the STFT hop size of $315 = 21 \times 15$ samples. 
Note that the stride is a factor of about 4.5 lower than the shortest kernel support (21 vs.\ 94).
For comparison, we have also experimented with smaller convolution strides (3 and 1) to assess their impact on the results.

For the `naive' fixed-size time-averaging variant standard average-pooling is used, implemented as a $15 \times 1$ 2D-pooling layer acting on the power spectrum. 
In this case the temporal averaging length is uniform over the frequency axis which is a crude approximation of the mathematical findings.
The `fixed-width' and `variable-width' cases are implemented using Hann windows, the latter with adaptive width, individual for each frequency bin. 
The maximum time support of this Hann window is 8 times the STFT hop size, equivalent to 
2520 samples.
The choice of Hann in contrast to a Boxcar window (as in the `naive' case) is motivated by its smoothness which aids adaptivity for the CNN training process.

\begin{figure}[]
\center{
\includegraphics[width=\columnwidth]{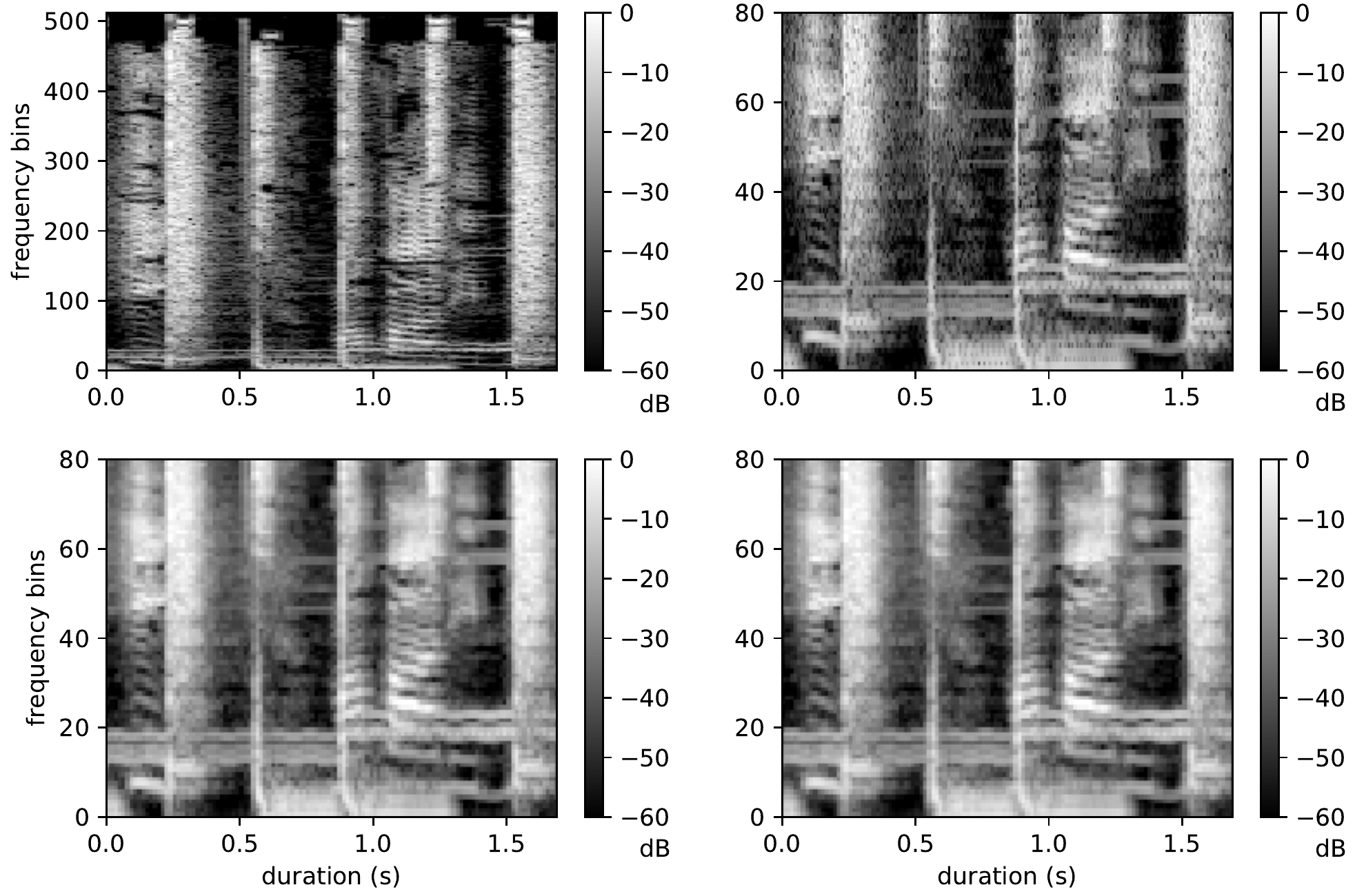}
\caption{Time-frequency representations for the problem of singing voice detection.
The spectrograms shown are STFT (upper left), STFT-based mel spectrogram (bottom left), 
filter bank computed (top right), and filter bank with time-averaging (bottom right).
}
\label{fig:tf_repr}}
\end{figure}

Figure~\ref{fig:tf_repr} illustrates the time-frequency representations used in this paper.
The STFT case is shown on the left-hand-side with the full Fourier spectrum (512 bins) on top and its mel-spectrogram (80 bins) at the bottom.
On the right-hand-side, the top shows a filter bank-computed mel-scaled spectrogram using the filters derived in Section~\ref{Sec:MelFil}, and the time-averaged counterpart at the bottom.
Note that the two bottom spectrograms are equivalent.


\subsection{Experimental results}
\label{ssec:exp_results}

\begin{figure}[]
\center{
\includegraphics[width=\columnwidth]{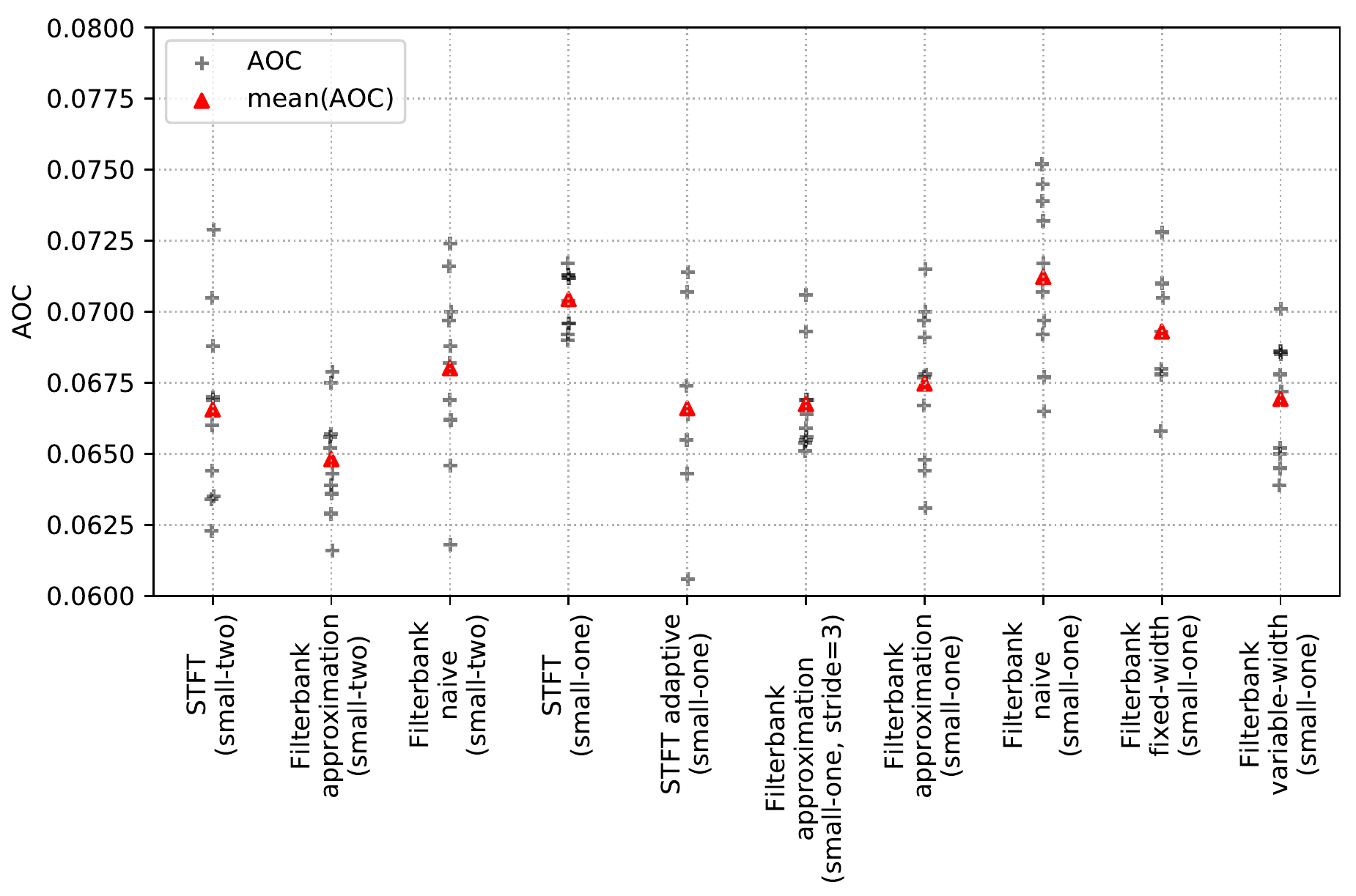}
\caption{AOC measures for the problem of singing voice detection.
Models compared are results for multiple runs of five-fold cross-validation on batch-normalized features, for the CNN architectures `small-two' and `small-one'.
On the one hand the features are mel-spectrograms 
(\emph{STFT}), or spectrograms with trained center frequencies (\emph{STFT adaptive}).
On the other hand, we evaluated
the approximative filters derived in Section~\ref{Sec:MelFil} (\emph{Filter bank approximation}),
filter banks and fixed-width temporal averaging (\emph{Filter bank naive} with a rectangular window, and \emph{Filter bank fixed-width} with a Hann window),
and adaptive variable-width Hann-window averaging (\emph{Filter bank variable-width}). 
The default convolution stride is 21, unless otherwise noted.
Shown are individual results (gray crosses) and their mean values (black dots).
}
\label{fig:ex_eval}}
\end{figure}

Figure~\ref{fig:ex_eval} shows the results of our CNN experiments for the problem of singing voice detection.
For our evaluations, we have switched from the simple error measure with the `optimal' (in the sense of maximum accuracy) threshold per experiment 
to the more informative `area over the ROC curve' measure (AOC), fusing classification errors for all possible thresholds into one measure.
A lower measure indicates a better result.

The reference implementation in \cite{schlueter:2015-ismir} uses pre-computed spectrograms, with a normalization globally on the the training set and eventual padding performed also on the spectrogram.
End-to-end learning as performed in our experiments demands on-line normalization (using a batch normalization layer) and padding directly on the audio time signal. 
We could verify that this yields a performance equivalent with the reference experiments.
 
We can also confirm that the performance of our `small-two' network with two classification layers is comparable to the large baseline architecture.
For AOC in the STFT case, the smaller networks score 6.66\% (`small-two') and 7.05\% (`small-one'), respectively, 
compared to 6.74\% of the original architecture (the latter not shown in Figure~\ref{fig:ex_eval}).
The difference between the reference and the 'small-two' architecture is not significant (t-test, $p = 5\%$), while the difference between 'small-two' and 'small-one' is.

%
%
%

In the course of experimentation it has become apparent that the time-averaging widths of the adaptive models hardly train at all,
especially for larger classification stages. 
They rather stay close to the initial values, while the CNN weights adapt instead. 
As a trick we have \emph{boosted} the widths' gradients for the back-propagation by a factor 3 to force the 
width parameters to adapt at a higher rate. Higher factors have proven unfeasible, causing the adaptation to run out of bounds.
%
Since the adaptation process is intricate, the choice of a starting value for the variable averaging length (time support of the Hann window) is important. 
We have tried values of 0.1, 0.2, 0.3, 0.5 of the maximum filter bank time support, with 0.2 (equivalent to 504 samples) leading to the best results.
\subsubsection{Interpretation of Results}
As a first observation, we see in Figure~\ref{fig:ex_eval} that for both architectures the filter bank approximation 
scores better than the canonical STFT case 
(significant for the 'small-one' architecture at $p < 5\%$). 
This can be explained by considering the different kinds of aliasing terms which affect
 the computation of the feature maps: Equation \eqref{eq:errboundaliases} shows that the
  STFT-based approach leads to aliasing in time 
while the filter bank-based approach leads to aliasing in  frequency.
 From the results we can deduce that the impact of the time-aliases imposed
  by mel-averaging is stronger than that of the frequency-aliases stemming from the time-averaging. 
  Furthermore, reducing time-aliases in the first approach would require using  a longer FFT
   in the computation of the underlying STFT, while reducing the influence of the frequency aliases 
   is accomplished by decreasing the convolution stride: using a default convolution 
   stride of $21$ corresponds to a subsampling factor $\alpha = 21/1024 = 0.02$
    in time as opposed to $\alpha = 315/1024 = 0.3$ in the STFT case. Heuristically, 
    we obtain a more stable estimate for the local frequency components, cp.~the 
    recent work in~\cite{ar17}.  We were able to confirm this trend by using even
     smaller convolution strides (subsampling factor $3$ instead of the standard $21$)
      which led to a slightly, albeit insignificantly, better score. These observations
       indicate that the actual time-frequency resolution of the signal representation 
       used in the first processing step can lead to advantages in the over-all 
       performance of the CNN, which cannot necessarily be provided  
       by subsequent convolutional or dense layers.  
       To our knowledge, this is the first formal description of such an effect.

The second observation concerns the influence of leveraging adaptivity in the learning process: In comparison to the filter banks with filter coefficient approximations according to the theory, the `naive' (significant at $p = 5\%$) and the `fixed-width' (not significant) variations
exhibit slightly worse performance for both architectures.
The `variable-width' variation with adaptive time-averaging scores significantly better than its 'fixed-width' counterpart, 
and is statistically equivalent ($p > 60\%$) to the filter bank `approximation' case.

At the same level of performance lies the `STFT adaptive' case which is a variation of the STFT case; here, we applied frequency averaging of the spectrogram coefficients, just as in the STFT-based computation of the mel-coefficients, but allowed the CNN to learn---and thus adapt---the center frequencies during training. The low and high frequency bounds remained fixed, but the intermediate frequencies were free to adapt with the condition of monotonicity.
It is noticeable that the adapted center frequencies remain relatively close to the Mel scale, with only a few percent of relative deviation ($Q_1 > -8.6\%$, $Q_3 < +3.7\%$ over all bands for 10 runs with 5 folds each).

In general, the different adaptive models exhibit very similar performance measures $\textrm{AOC}<6.75\%$ which is significantly better than the canonical STFT-based case at $\textrm{AOC}=7.05\%$ for the `small-one' architecture. As expected, the effects of adaptivity on the evaluation results are more pronounced for the smaller architecture, with less explanatory power in the classification stages.
We remark that in previously performed experiments on a fully adaptive 
approach (adaptive filter lengths + adaptive time averaging) the learning process did not  converge and the achieved 
 results were consistently worse than those based on  standard approaches. 
Therefore, we restricted the relaxation of fixed parameters to either time-averaging length or 
frequency centers  in the computation of the  now  variable, adaptive frequency 
filters and the experiments showed  that both approaches perform almost identically. In these scenarios, only 80 trainable parameters are added to the number of networks weights described in Section~\ref{Sec:TP} and the increase in computational cost
as well as required number of training data is negligible. 

Finally, note that the filter bank approximation with stride 3 in the convolution performs identically
to the setup with the adaptive time-averaging (variable-width), which, in turn, is slightly better 
than the stride 21 approximation setting. The fact that the improvement is only small, can be 
seen as an indication that the ideal mel-coefficients indeed yield a representation that is 
sufficiently good for the subsequent convolutional layers to get close to an optimal result. 
Furthermore, as stated before, it seems that the expressivity of the network architecture is 
so high that it can actually obtain good results from different representations which are sufficiently 
reasonable. In this sense, the observation that some adaptivity in the primary representation on the one hand 
and the geometry of the sampling grid and consequential nature of occurring aliases
do have some influence on the final performance, is quite remarkable.

\section{Discussion and perspectives}
\label{Sec:Disc}  
In Section~\ref{Sec:Questions} we posed two questions concerning the application of alternative time-frequency representations for learning problems in music information retrieval. 

First, it has been analytically shown under which conditions mel-spectrogram coefficients can be reproduced 
by applying frequency-adaptive filters followed by time-averaging the squared amplitudes.
In practice, this procedure will always lead to approximate values due to their computation from sub-sampled values. 

Answering the second question, we have found that these \emph{designed} spectrogram representations yield significantly increased performance on the task of CNN-based singing voice detection. The improvement in performance can be ascribed to a sub-sampling scheme implicit in the usage of the designed adaptive filters, which yields a more advantageous suppression of adversarial time-frequency aliases than the canonical computation of mel-spectrogram coefficients. 
Furthermore, adaptivity by \emph{training} in the time-averaging layer, or alternatively, using frequency-adaptive triangular filters on the Fourier spectrograms, on the other hand, 
also lead to improved results relative to the canonical STFT-based mel-spectrograms.
These results are performance-wise statistically equivalent to the filters derived by the mathematical theory developed in Section~\ref{Sec:MelFil}. Hence, similar results were obtained both with properly \emph{designed} representations and representations whose crucial parameters were  \emph{trained} on the data. 

Summing up, we conclude that the subtle differences in time-frequency resolution of the basic filters used to obtain the signal representation do influence the over-all performance of a CNN applied to a typical MIR task, at least for architectures of rather modest size.
The choice of the well-established mel frequency scale in the first place for our experiments seems justified
not only by prior work on time domain filters calculated \emph{ex nihilo} (cf.~\cite[Section 4.2]{disc14}),
but also by our own findings that adaptive center frequencies deviate from the Mel scale only to a small extent.
We conclude that the chosen scale provides a useful compromise between time and frequency averaging 
for the task under consideration.

Future work on the problem of learned basic filters in MIR tasks will involve the  study of the precise connection between the characteristics of a given data set and  the most advantageous analysis windows and sampling schemes used to compute the spectrogram.  These investigations will concern both  the network's expressivity and the performance of the learning process, cf.~preliminary work in \cite{badogr17} and will be based on data sets with different time-frequency characteristics as well as various learning tasks. Finally, future work will also address the more general question of the propagation and alleviation of small approximation errors through the network and their dependence on various network parameters as well as the network's architecture, relying on existing results on stability of CNNs, compare~\cite{ICML2016,bado17,SIT2016}.
%
%
%
%
\begin{appendix}
\section{Proof of Theorem~\ref{Th:subsampled case}}\label{AppA}
In order to include the situation described in Theorem~\ref{Th:subsampled case}, we assume the situation in which the original spectrogram is sub-sampled, in other words, we start  the computations concerning a signal $f$ from 
$$S_0 ( \alpha l, \beta k)= |\mathcal{V} _g f (\alpha l, \beta k) |^2 = |\mathcal{F} (f\cdot T_{\alpha l} g)(\beta k)|^2.$$
The proof 
is based on the observation that the mel-spectrogram can be written via the operation of so-called {\it STFT- or Gabor-multipliers},  cf.~\cite{feno03}, on any given function in the sense of a bilinear form. Before deriving the involved correspondence, we thus introduce this important class of operators.

Given a window function $g$, time- and frequency-sub-sampling parameters $\alpha, \beta$, respectively, and a function $\bf{m}: \mathbb{Z} \times \mathbb{Z} \mapsto \mathbb{C} $, the corresponding Gabor multipler $G^{\alpha,\beta}_{g, \bf{m}}$ is defined as
\begin{equation*}
G^{\alpha,\beta}_{g, \bf{m}} f = \sum_k \sum_l  {\bf m} (k,l) \langle f, M_{\beta k} T_{\alpha l} g\rangle  M_{\beta k} T_{\alpha l} g .
\end{equation*}

We next derive the expression of a mel-spectrogram by an appropriately chosen  Gabor multiplier. Using sub-sampling factors $\alpha$ in time and $\beta$ in frequency as before,   we start from \eqref{eq:melspec} and reformulate as follows: 
\begin{align}\notag
	\operatorname{MS}_{g}(f) (b,\nu )=& \sum_k |\mathcal{F} (f\cdot T_b g)(\beta k)|^2 \cdot \Lambda_\nu (\beta k)\\\notag
	= & \sum_k \langle f, M_{\beta k} T_b g\rangle \overline{\langle f, M_{\beta k} T_b g\rangle} \Lambda_\nu (\beta k)\\\notag
	= & \langle  \sum_k \Lambda_\nu ( \beta k) \langle f, M_{\beta k} T_b g\rangle M_{\beta k} T_b g , f\rangle\\\notag
			= & \langle  \sum_k \sum_l  {\bf m} (k,l) \langle f, M_{\beta k} T_{\alpha l} g\rangle M_{\beta k} T_{\alpha l} g , f\rangle
\end{align}
with $ {\bf m} (k,l)  = \delta(\alpha l-b)\Lambda_\nu (\beta  k)$.  We see that the mel-coefficients can thus be interpreted via a Gabor multiplier: $ \operatorname{MS}_{g}(f) (b,\nu ) = \langle G^{\alpha,\beta}_{g, \bf{m}}f, f \rangle$.

The next step is to switch to an alternative operator representation. Indeed, as shown in \cite{feko98},
every  operator $H$ can equally be written by means of its {\it spreading function} $\eta_H$ as
\begin{equation}\label{eq:SF}
Hf (t) = \int_x \int_\xi \eta_H (x,\xi)  f (t-x) e^{2\pi i t \xi}d\xi dx.
\end{equation}
We note that two operators $H_1$, $H_2$ are equal if and only if  their spreading functions coincide, see \cite{doto10,feko98} for details.

As shown in \cite{doto10}, a  Gabor multiplier's spreading function $\eta^{\alpha,\beta}_{{g, {\bf m}} }$  is given by
\begin{equation}\label{eq:Gmspread}
\eta^{\alpha,\beta}_{{g, {\bf m}} } (x,\xi) = \mathcal{M} (x,\xi) \mathcal{V} _g g(x,\xi), 
\end{equation}
where $\mathcal{M} (x,\xi)$ denotes the $(\beta^{-1}, \alpha^{-1})$-periodic symplectic Fourier transform of ${\bf m}$, i.e.,
\begin{equation}\label{eq:sympFT}
\mathcal{M} (x,\xi) = \cF_s ( {\bf m} )(x,\xi)  = \sum_k\sum_l  {\bf m} (k,l)  e^{-2\pi i (\alpha l \xi - \beta kx )}.
\end{equation}
We now equally rewrite the time-averaging operation applied to a filtered signal, as defined in~\eqref{eq:FFBdef},  as a Gabor 
multiplier.
As before, we set $\check{h}_\nu (t) = \overline{h_\nu (-t)} $ and have
\begin{align}\notag
\operatorname{FB}_{h_\nu}(f) (b,\nu ) =\sum_l |(f\ast h_\nu)(\alpha l)|^2 \cdot \varpi_\nu (\alpha l-b) &= 
\sum_l |\sum_n f(n) \check{h}_\nu(n-\alpha l)|^2 \cdot \varpi_\nu (\alpha l-b)\\\notag
=\sum_k \sum_l |\langle f, M_{\beta k} T_{\alpha l}  \check{h}_\nu \rangle |^2 \cdot\varpi_\nu (\alpha l-b)\delta(\beta k)&= \langle G^{\alpha ,\beta}_{\check{h}_{\nu }, {\bf m}_F} f, f\rangle.
\end{align}
with ${\bf m}_F (k,l)  = T_b \varpi_\nu (l) \delta(\beta k)$.
To obtain the error estimate in Corollary~\ref{Th:subsampled case}, first note that, by straight-forward computation using the operators' representation by their spreading functions as in \eqref{eq:SF}
\begin{eqnarray*}
|\operatorname{MS}_{g}(f) (b,\nu )-\operatorname{FB}_{h_\nu}(f)(b,\nu )| =&
 |\langle (G^{\alpha,\beta}_{g, {\bf m}} - G^{\alpha ,\beta}_{\check{h}_{\nu }, {\bf m}_F}) f, f\rangle | \\\notag
= |\langle (\eta_{g_\alpha^\beta, {\bf m}} - \eta_{\check{h}_{\alpha \nu }^\beta, {\bf m}_F}), \mathcal{V}_f f\rangle | 
\leq& \|\eta^{\alpha, \beta}_{g, {\bf m}} - \eta^{\alpha, \beta}_{\check{h}_{ \nu } ,{\bf m}_F}\|\cdot \| f\|_2^2
\end{eqnarray*}
and we can estimate the error by the difference of the spreading functions. 
We
write the sampled version of $\Lambda_\nu  $ by using the Dirac comb
  $\sha_\beta$:  $\Lambda_\nu (\beta k) =  (\sha_\beta \Lambda_\nu) (t)  = \sum_k \Lambda_\nu (t) \delta (t-\beta k) $ 
  and analogously for  $ \varpi_\nu $ using $\sha_\alpha$  to obtain
 $ {\bf m}  =T_b \delta (\alpha l) \cdot  \sha_\beta \Lambda_\nu  $ and ${\bf m}_F = \sha_\alpha T_b \varpi_\nu  \cdot \delta (\beta k)$. 
Applying  the symplectic Fourier transform \eqref{eq:sympFT}  to $ {\bf m} $ then gives: 
\begin{eqnarray*}
\mathcal{M}^\nu (x,\xi) &=&  \sum_k\sum_l  {\bf m} (k,l)  e^{-2\pi i (\alpha l \xi - \beta kx )}\\
&= &\int_t  \sum_k \Lambda_\nu (t) \delta (t-\beta k)e^{2\pi i t x} dt \sum_l  T_b \delta (\alpha l) e^{-2\pi i \alpha l \xi }\\
&=& \cF^{-1} (\sha_\beta\Lambda_\nu )(x)  \cdot  e^{-2\pi i b \xi} 
\end{eqnarray*}

Now it is a well-known fact that the Fourier transform 
  turns sampling with sampling interval $\beta$ into periodization by $1/\beta$, in other words, into a convolution with 
  $\sha_{\frac{1}{\beta}}$:
\[\cF^{-1} (\sha_\beta  \Lambda_\nu )(x) =\sha_{\frac{1}{\beta}}\ast \cF^{-1} (\Lambda_\nu )(x)  = \sum_l T_{\frac{l}{\beta} }\mathcal{F}^{-1} (\Lambda_\nu ) (x),\]
hence
\[ \mathcal{M}^\nu (x,\xi)  = \sum_l T_{\frac{l}{\beta} }\mathcal{F}^{-1} (\Lambda_\nu ) (x)  \cdot  e^{-2\pi i b \xi}. \]
 Completely analogous considerations for  $ \varpi_\nu $  and $\sha_\alpha$ lead to the periodization of  $\cF(\varpi_\nu)$ and thus the following expression for the symplectic Fourier transform of  ${\bf m}_F$: 
 \[\mathcal{M}^\nu_F(x,\xi)  = \sum_l T_{\frac{l}{\alpha } }\mathcal{F}(\varpi_\nu) (\xi)  \cdot  e^{-2\pi i b \xi}. \]

Plugging these expressions into \eqref{eq:Gmspread} gives the bound \eqref{eq:errboundaliases}.

\begin{remark}
It is interesting to interpret the action of an operator in terms of its spreading function. In view of \eqref{eq:SF}, we see that the spreading function determines the amount of shift in time and frequency, which the action of the operator imposes on  a function. 
For Gabor multipliers, if well-concentrated window functions are used, it is immediately obvious that the amount of shifting is moderate as well as determined by the window's eccentricity. At the same time, the aliasing effects introduced by coarse sub-sampling are  reflected in the periodic nature of $\mathcal{M}$.  Since, for $\cF^{-1} (\Lambda_\nu )$ the sub-sampling density in frequency, determined by $\beta$, and for  $\cF(\varpi_\nu)$ the sub-sampling density in time, determined by $\alpha$, determine the amount of aliasing, the over-all approximation quality deteriorates with increasing sub-sampling factors.
\end{remark}
\end{appendix}

\section*{Conflict of Interest}
The authors declare that they have no conflict of interest.

\bibliographystyle{spmpsci}
\iffalse
\bibliography{references}

\begin{thebibliography}{10}
\providecommand{\url}[1]{{#1}}
\providecommand{\urlprefix}{URL }
\expandafter\ifx\csname urlstyle\endcsname\relax
  \providecommand{\doi}[1]{DOI~\discretionary{}{}{}#1}\else
  \providecommand{\doi}{DOI~\discretionary{}{}{}\begingroup
  \urlstyle{rm}\Url}\fi

\bibitem{ar17}
Abreu, L. D., Romero, J. L.: {MSE} estimates for multitaper spectral estimation and
  off-grid compressive sensing.
\newblock IEEE Trans. Inf. Theory. \textbf{63}(12), 7770--7776 (2017)

\bibitem{anma14}
And\'{e}n, J., Mallat, S.: Deep scattering spectrum.
\newblock IEEE Transactions on Signal Processing \textbf{62}(16), 4114--4128
  (2014)

\bibitem{an13}
Anselmi, F., Leibo, J.Z., Rosasco, L., Mutch, J., Tacchetti, A., Poggio, T.A.:
  Unsupervised learning of invariant representations in hierarchical
  architectures.
\newblock CoRR \textbf{abs/1311.4158} (2013).
\newblock \urlprefix\url{http://arxiv.org/abs/1311.4158}

\bibitem{badokoto13}
Balazs, P., D{\"o}rfler, M., Kowalski, M., Torr{\'e}sani, B.: Adapted and adaptive
  linear time-frequency representations: a synthesis point of view.
\newblock IEEE Signal Processing Magazine \textbf{30}(6), 20--31 (2013)

\bibitem{badohojave11}
Balazs, P., D{\"o}rfler, M., Jaillet, F., Holighaus, N., Velasco, G.: Theory,
  implementation and applications of nonstationary gabor frames.
\newblock Journal of Computational and Applied Mathematics \textbf{236}(6),
  1481--1496 (2011)

\bibitem{bado17}
Bammer, R., D{\"o}rfler, M.: Invariance and stability of Gabor scattering for
  music signals.
\newblock In: Sampling Theory and Applications (SampTA), 2017 International
  Conference on, pp. 299--302. IEEE (2017)

\bibitem{disc14}
Dieleman, S., Schrauwen, B.: End-to-end learning for music audio.
\newblock In: Acoustics, Speech and Signal Processing (ICASSP), 2014 IEEE
  International Conference on, pp. 6964--6968 (2014).
\newblock \doi{10.1109/ICASSP.2014.6854950}

\bibitem{do01}
D{\"o}rfler, M.: Time-frequency analysis for music signals: A mathematical
  approach.
\newblock Journal of New Music Research \textbf{30}(1), 3--12 (2001)

\bibitem{badogr17}
D{\"o}rfler, M., Bammer, R., Grill, T.: Inside the spectrogram: Convolutional
  neural networks in audio processing.
\newblock In: International Conference on Sampling Theory and Applications
  (SampTA), pp. 152--155. IEEE (2017)

\bibitem{doto10}
{D}{\"o}rfler, M., {T}orr{\'e}sani, B.: {R}epresentation of operators in the
  time-frequency domain and generalized {G}abor multipliers.
\newblock {J}. {F}ourier {A}nal. {A}ppl. \textbf{16}(2), 261--293 (2010)

\bibitem{feko98}
{F}eichtinger, H.G., {K}ozek, W.: {Q}uantization of {T}{F} lattice-invariant
  operators on elementary {L}{C}{A} groups.
\newblock In: H.G. {F}eichtinger, T.~{S}trohmer (eds.) {G}abor analysis and
  algorithms, {A}ppl. {N}umer. {H}armon. {A}nal., pp. 233--266.
  {B}irkh{\"a}user {B}oston (1998)

\bibitem{feno03}
{F}eichtinger, H.G., {N}owak, K.: {A} first survey of {G}abor multipliers.
\newblock In: H.G. {F}eichtinger, T.~{S}trohmer (eds.) {A}dvances in {G}abor
  {A}nalysis, {A}ppl. {N}umer. {H}armon. {A}nal., pp. 99--128. {B}irkh{\"a}user
  (2003)

\bibitem{Goodfellow-et-al-2016}
Goodfellow, I., Bengio, Y., Courville, A.: Deep learning.
\newblock MIT Press (2016)

\bibitem{grsc15}
Grill, T., Schl{\"u}ter, J.: {Music Boundary Detection Using Neural Networks on
  Combined Features and Two-Level Annotations}.
\newblock In: Proceedings of the 16th International Society for Music
  Information Retrieval Conference (ISMIR 2015). Malaga, Spain (2015)

\bibitem{SIT2016}
Grohs, P., Wiatowski, T., B{\"o}lcskei, H.: Deep convolutional neural networks
  on cartoon functions.
\newblock In: Information Theory (ISIT), 2016 IEEE International Symposium on,
  pp. 1163--1167. IEEE (2016)

\bibitem{dogrhove13}
{H}olighaus, N., {D}{\"o}rfler, M., {V}elasco, G.A., {G}rill, T.: {A} framework
  for invertible, real-time constant-{Q} transforms.
\newblock {I}{E}{E}{E} {T}rans. {A}udio {S}peech {L}ang. {P}rocess.
  \textbf{21}(4), 775 --785 (2013)

\bibitem{adam15}
Kingma, D., Ba, J.: Adam: A method for stochastic optimization.
\newblock In: Proceedings of the 6th International Conference on Learning
  Representations (ICLR). San Diego, USA (2015)

\bibitem{Lecun98}
LeCun, Y., Bottou, L., Bengio, Y., Haffner, P.: Gradient-based learning applied
  to document recognition.
\newblock Proceedings of the IEEE \textbf{86}(11), 2278--2324 (1998)

\bibitem{mallat}
Mallat, S.: {Group Invariant Scattering}.
\newblock Comm. Pure Appl. Math. \textbf{65}(10), 1331--1398 (2012)

\bibitem{ma16}
Mallat, S.: Understanding deep convolutional networks.
\newblock Philosophical Transactions of the Royal Society of London A:
  Mathematical, Physical and Engineering Sciences \textbf{374}(2065) (2016).
\newblock \doi{10.1098/rsta.2015.0203}.
\newblock
  \urlprefix\url{http://rsta.royalsocietypublishing.org/content/374/2065/20150203}

\bibitem{schlueter:2015-ismir}
Schl{\"u}ter, J., Grill, T.: {Exploring Data Augmentation for Improved Singing
  Voice Detection with Neural Networks}.
\newblock In: Proceedings of the 16th International Society for Music
  Information Retrieval Conference (ISMIR 2015). Malaga, Spain (2015)

\bibitem{wigrbo17}
Wiatowski, T., Grohs, P., B{\"o}lcskei, H.: Energy propagation in deep
  convolutional neural networks.
\newblock arXiv preprint arXiv:1704.03636  (2017)

\bibitem{ICML2016}
Wiatowski, T., Tschannen, M., Stanic, A., Grohs, P., B{\"o}lcskei, H.: Discrete
  deep feature extraction: A theory and new architectures.
\newblock In: Proceedings of the International Conference on Machine Learning,
  pp. 2149--2158 (2016)
  
  \bibitem{Wa17}  Waldspurger, I.: Exponential decay of scattering coefficients. 
  \newblock In: Proceedings of 2017 International Conference on Sampling Theory and Applications (SampTA), IEEE,
  2017.
   \bibitem{Wa15}  Waldspurger, I.: Wavelet transform modulus: phase retrieval and scattering.
  \newblock PhD thesis, \'{E}cole Normale Sup\'{e}rieure, Paris, 2015.

\end{thebibliography}
\else

\fi
\end{document}